\documentclass[journal]{IEEEtran}
\usepackage{graphicx}
\usepackage{amsmath, amssymb}
\usepackage{caption, subcaption}
\usepackage{multirow}
\usepackage{soul,color}
\usepackage{cite}
\begin{document}
\title{Pyramid Hierarchical Transformer for Hyperspectral Image Classification}
\author{Muhammad Ahmad, Muhammad Hassaan Farooq Butt, Manuel Mazzara, Salvatore Distifano
\thanks{M. Ahmad is with the Department of Computer Science, National University of Computer and Emerging Sciences, Islamabad, Chiniot-Faisalabad Campus, Chiniot 35400, Pakistan, and Dipartimento di Matematica e Informatica---MIFT, University of Messina, Messina 98121, Italy; (e-mail: mahmad00@gmail.com)}
\thanks{M. H. F. Butt is with the Institute of Artificial Intelligence, School of Mechanical and Electrical Engineering, Shaoxing University, Shaoxing 312000, China. (e-mail: hassaanbutt67@gmail.com}
\thanks{M. Mazzara is with the Institute of Software Development and Engineering, Innopolis University, Innopolis, 420500, Russia. (e-mail: m.mazzara@innopolis.ru)}
\thanks{S. Distefano is with  Dipartimento di Matematica e Informatica---MIFT, University of Messina, Messina 98121, Italy. (e-mail: sdistefano@unime.it)}
}
\markboth{Journal of \LaTeX\ Class Files,}%
{Ahmad \MakeLowercase{\textit{et al.}}: Bare Demo of IEEEtran.cls for IEEE Journals}
\maketitle
\begin{abstract}
The traditional Transformer model encounters challenges with variable-length input sequences, particularly in Hyperspectral Image Classification (HSIC), leading to efficiency and scalability concerns. To overcome this, we propose a pyramid-based hierarchical transformer (PyFormer). This innovative approach organizes input data hierarchically into segments, each representing distinct abstraction levels, thereby enhancing processing efficiency for lengthy sequences. At each level, a dedicated transformer module is applied, effectively capturing both local and global context. Spatial and spectral information flow within the hierarchy facilitates communication and abstraction propagation. Integration of outputs from different levels culminates in the final input representation. Experimental results underscore the superiority of the proposed method over traditional approaches. Additionally, the incorporation of disjoint samples augments robustness and reliability, thereby highlighting the potential of our approach in advancing HSIC. 

The source code is available at https://github.com/mahmad00/PyFormer.
\end{abstract}
\IEEEpeerreviewmaketitle
\section{Introduction}

\IEEEPARstart{H}{yperspectral Image Classification (HSIC)} is crucial in diverse domains, including remote sensing \cite{ahmad2021hyperspectral}, earth observation \cite{lodhi2018hyperspectral}, urban planning \cite{li2024HD}, agriculture \cite{lu2020recent}, forestry \cite{adao2017hyperspectral}, target/object detection \cite{li2023lrr}, mineral exploration \cite{bedini2017use}, environmental monitoring \cite{weber2018hyperspectral, stuart2019hyperspectral}, climate change \cite{pande2023application}, food processing \cite{khan2021hyperspectral, khan2020hyperspectral}, bakery products \cite{saleem2020prediction}, bloodstain identification \cite{butt2022fast, zulfiqar2021hyperspectral}, and meat processing \cite{ayaz2020hyperspectral, ayaz2020myoglobin}. The abundance of spectral data in Hyperspectral Images (HSIs) presents challenges and opportunities for classification \cite{hong2024spectralgpt}. While CNNs \cite{ahmad2020fast, 10423094} and its variants \cite{10433668, 9170817, 10409250}, and Transformers \cite{yao2023extended, 10399798, 9868046} have shown success in computer vision tasks, there is a growing interest in exploring Transformer models for advancing HSI analysis.

The vision and spatial-spectral transformers (SSTs) \cite{10423821, rs13030498, 10409287, rs16020404, 10432978, 10422823, 10399888, 10387229} excel in capturing global contextual information via self-attention mechanisms, facilitating simultaneous consideration of relationships between all HSI regions \cite{9895238}. Unlike CNNs, SSTs demonstrate strong scalability to high-resolution HSIs, effectively handling large datasets without complex pooling operations. Their adaptability contributes to widespread applicability in HSIC. Furthermore, SSTs learn stratified representations directly from raw pixel values, simplifying the model-building process and often leading to improved performance \cite{9627165}. The interpretability of attention maps generated by SSTs aids in understanding model decision-making, highlighting influential image regions \cite{10387571}.

Despite their success, SSTs have limitations, for instance, training large SSTs can be computationally demanding \cite{10419133, 10379170}. The self-attention mechanism introduces quadratic complexity with respect to sequence length, potentially hindering scalability \cite{HUANG2024109897}. Unlike CNNs, which inherently possess translation invariance, SSTs may struggle to capture spatial relationships invariant to small translations in the input \cite{SUN2024102163}. Moreover, the tokenization process of dividing input images into fixed-size patches may not efficiently capture fine-grained details \cite{10418237, 10400415}. The quadratic scaling of self-attention poses challenges, particularly with long sequences. Furthermore, optimal performance often requires substantial training data, and training on smaller datasets may lead to overfitting, limiting effectiveness in scenarios with limited labeled data \cite{10399798}.

Therefore, this work proposed a pyramid hierarchical SST for HSIC. The hierarchical structure partitions the input into segmentation, each denoting varying abstraction levels, organized in a pyramid-like manner. Transformer modules are applied at each level for multi-level processing, ensuring efficient capture of local and global context. Information flow occurs both spatially and spectrally within the hierarchy, fostering communication and abstraction propagation. Integration of Transformer outputs from different levels yields the final output maps. In short, the following contributions are made;

\begin{enumerate}
    \item \textbf{Hierarchical Segmentation:} Input sequences are divided into hierarchical segments, each representing varying levels of abstraction or granularity.

    \item \textbf{Pyramid Organization:} These segments adopt a pyramid-like structure, wherein the lowest level retains detailed information while higher levels convey increasingly abstract representations.

    \item \textbf{Multi-level Processing:} Transformer modules are independently applied at each level of the hierarchy, facilitating efficient capture of both local and global context.
\end{enumerate}

\begin{figure*}[!hbt]
    \centering
    \includegraphics[width=0.80\textwidth]{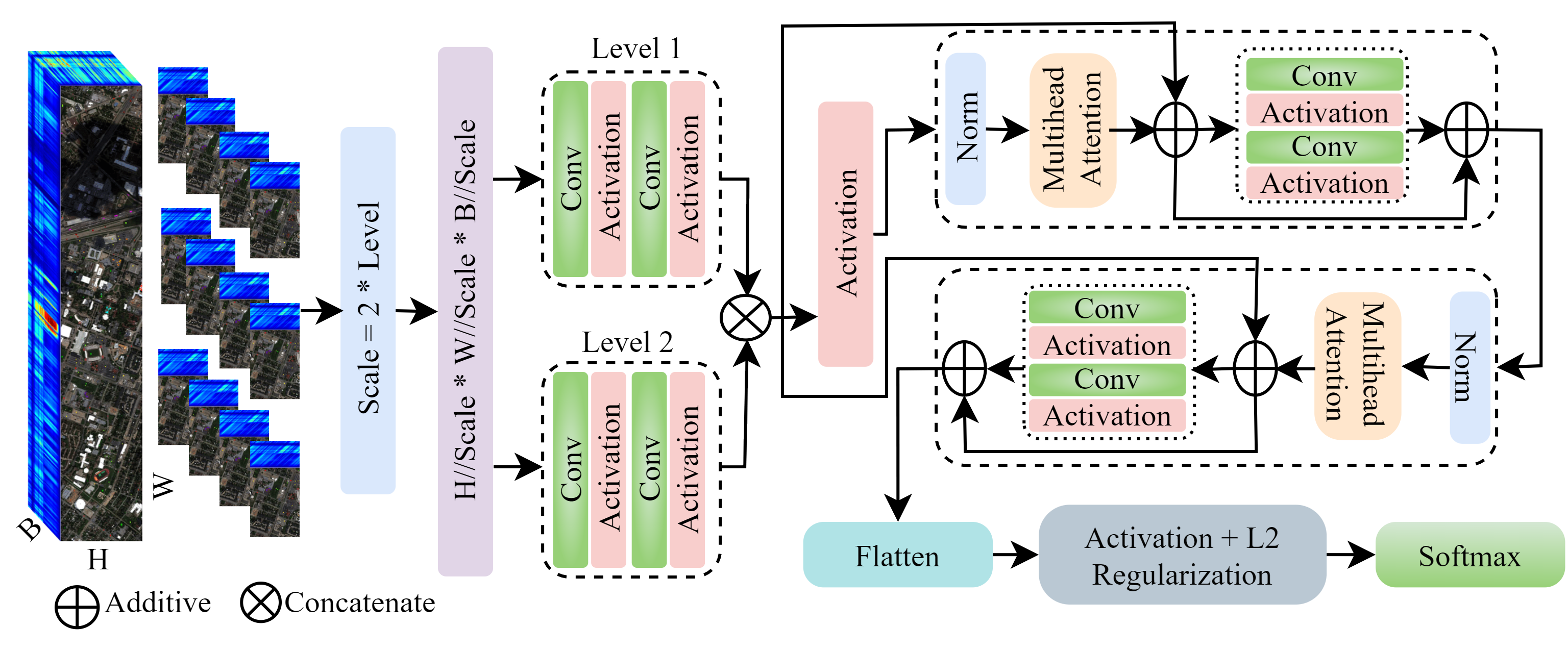}
    \caption{The Pyramid-type Hierarchical Transformer features a structured pyramid block comprising two levels. Each level incorporates two convolutional layers with filter sizes of ($32 \times S \times S \times B$) and ($64 \times S \times S \times B$), respectively, followed by a residual connection and activation function. The hierarchical transformer block, receiving the learned multi-scale information, consists of two layers and four multi-heads. The acquired information undergoes flattening and is subsequently subjected to the ReLU activation function and L2 regularization technique. This regularization aids in mitigating overfitting by reducing weights, rendering the network less responsive to minor input variations. Finally, the output layer employs softmax activation for final maps.}
    \label{Fig1}
\end{figure*}

\section{Proposed Methodology}

An HSI cube, denoted as $X = \{x_i, y_i\} \in \mathcal{R}^{(M \times N \times B)}$, comprises spectral vectors $x_i = \{x_{i,1}, x_{i,2}, x_{i,3}, \dots, x_{i,L}\}$, and $y_i$ and corresponding class labels $x_i$. The cube is initially divided into overlapping 3D patches, each centered at spatial coordinates $(\alpha, \beta)$ and spanning $S \times S$ pixels across $B$ bands. The total count of extracted patches ($m$) from $X$ is $(M-S+1) \times (N-S+1)$, where a patch $P_{\alpha, \beta}$ covers spatial dimensions within $\alpha \pm \frac{S-1}{2}$ and $\beta \pm \frac{S-1}{2}$. These patches, along with their central pixel labels, constitute the training  $X_{train}$, validation $X_{val}$, and a test $X_{test}$ sets, ensuring $|X_{train}| \ll |X_{val}|$ and $|X_{train}| \ll |X_{test}|$ to prevent sample overlaps and biases. The model's overall structure is presented in Figure \ref{Fig1}. 

Let ($S, S, B^*$) denote the input shape, $B^*$ where represents the reduced number of bands. The scaling factor, $Scale = 2^{Level}$, is computed. This scale is utilized to derive the input shape for the pyramid structure layers, given by $Input~shape = (\frac{S}{Scale}, \frac{S}{Scale}, \frac{B^*}{Scale})$. This input is fed into a convolutional layer to extract spatial-spectral semantic features from HSI patches. Each patch, with dimensions ($S \times S \times B^*$), undergoes processing using 3D convolutional layers with kernel sizes ($32 \times 1 \times S \times S$) and ($64 \times 1 \times S \times S$), along with ReLU activation. The activation maps for the spatial-spectral position $(x, y, z)$ at the i-th feature map and j-th layer are denoted as $f_{i,j}^{(x, y, z)}$ \cite{ahmad2020fast};

\begin{multline}
f_{(i,j)}(x,y,z) = \sigma\bigg(\sum_{m=1}^M \sum_{n=1}^N \sum_{p=1}^P w_{(i,j, m, n, p)} \times \\
    x_{(m+x-1, n+y-1, p+z-1) + b_{i,j}}\bigg)
\end{multline}
Where $f_{i,j}(x, y, z)$ is the activation value, $\sigma$ is the activation function, $w_{i,j,m,n,p}$ are the weights of the convolutional kernel, $x_{m+x-1, n+y-1, p+z-1}$ represents the input patch values, and $b_{i,j}$ is the bias term. Let $f_{i,j} \in \mathbb{R}^{N \times D}$ denote the input tensor to the transformer, where $N$ represents the number of patches, and $D$ signifies the dimensionality of each patch after the convolutional process. This encoding is integrated with the input embeddings, augmenting the model with spatial arrangement details. The foundational architecture of the transformer centers around the encoder, which consists of multiple layers incorporating multimodal attention and feedforward convolutional networks. The attention mechanism assumes a critical role in enabling the model to capture intricate relationships between distinct patches. Specifically, for a given input $v_{i,j}$ each layer within the transformer encoder encompasses; 

\begin{equation}
    H_{att}^{l} = Attention(H^{(l-1)})
    \label{eq2}
\end{equation}

\begin{equation}
    H_{FF\_Conv}^{l} = FF\_Conv(H_{att}^{l})
    \label{eq3}
\end{equation}

\begin{equation}
    H^{l} = H^{(l-1)} + H_{FF\_Conv}^{l}
    \label{eq4}
\end{equation}

Equations \ref{eq2}, \ref{eq3}, and \ref{eq4} define the attention, feedforward convolutional neural network, and residual connections within the transformer layers, respectively. Later, $H^{l}$ is flattened as $W = reshape(H^{l}, (m \times n \times p, 1))$, where $m, n$ and $p$ represent the batch size, height, and width, respectively. Subsequently, $L_2$ regularization $L_2Reg = \lambda|W|^2_2$ is added to the loss function during training, with ReLU as the activation function and $\lambda = 0.01$ as the regularization parameter. Finally, a softmax function is employed to generate the classification maps.

\section{Experimental Results and Discussion}

To ensure fairness, maintaining identical experimental conditions is crucial, including consistent geographical locations for model selection and uniform sample numbers for each training round in cross-validation. Random sample selection can introduce variability, potentially causing discrepancies among models executed at different times. Another common issue in recent literature is the overlap of training and test samples, leading to biased models with inflated accuracy. To mitigate this, the proposed method ensures that while training, validation, and test samples are randomly selected, efforts are made to prevent any overlap between these sets, thereby reducing biases introduced by overlapping samples. In our experimental setup, the proposed PyFormer was assessed using a mini-batch size of 128, the Adam optimizer, and specific learning parameters i.e., learning rate of 0.0001 and a decay rate of 1e-06, over 50 epochs. We systematically tested various configurations to comprehensively evaluate the proposed model. This exploration aimed to thoroughly understand the model's performance under diverse training scenarios and spatial resolutions.

\begin{table}[!hbt]
\centering
\caption{Performance on different train ratios across different datasets. Figure \ref{Fig3} illustrates the geographical maps showcasing the best results attained in these experiments.}
\label{Tab2}
\begin{tabular}{cccccccc} \hline
\multirow{2}{*}{\textbf{Datasets}} & \multirow{2}{*}{\textbf{Metrics}} & \multicolumn{5}{c}{\textbf{Data Split Ratio}} \\ \cline{3-7} 
&  & \textbf{5\%} & \textbf{10\%} & \textbf{15\%} & \textbf{20\%} &\textbf{25\%} \\ \hline
\multirow{4}{*}{PU} & OA & 96.28 & 98.33 & 99.35 & 98.02 & \textbf{99.80} \\ 
 & AA & 93.81  & 97.08 & 98.83 & 97.35 & \textbf{99.63} \\ 
 & KA & 95.07 & 97.78 & 99.14 & 97.39 & \textbf{99.73} \\
 & F1 & 94.44 & 97.33 &99.0  & 96.55 & \textbf{99.66} \\ \hline
\multirow{4}{*}{SA} & OA & 97.53 & 99.08 &99.33  & 99.75 & \textbf{99.86} \\ 
 & AA & 98.37 &  94.42& 99.71 & 99.80 & \textbf{99.83} \\ 
 & KA & 97.25 &98.98  & 99.27&  99.72& \textbf{99.84} \\
 & F1 & 98.43 & 99.31 & 99.81 &99.87  & \textbf{100.0} \\ \hline
 \multirow{4}{*}{HU} & OA & 93.11 & 97.36 & 97.13 & \textbf{98.19} & 98.18\\ 
 & AA & 92.11 & 95.97 & 96.46 & 97.28 & \textbf{98.26} \\ 
 & KA & 92.55 & 97.14 & 96.9 & \textbf{98.05} & 98.03 \\
 & F1 & 86.56 & 90.68 & 90.56 & 91.62 & \textbf{92.25} \\ \hline
\end{tabular}
\end{table}
\begin{figure}[!hbt]
    \centering
	\begin{subfigure}{0.24\textwidth}
		\includegraphics[width=0.99\textwidth]{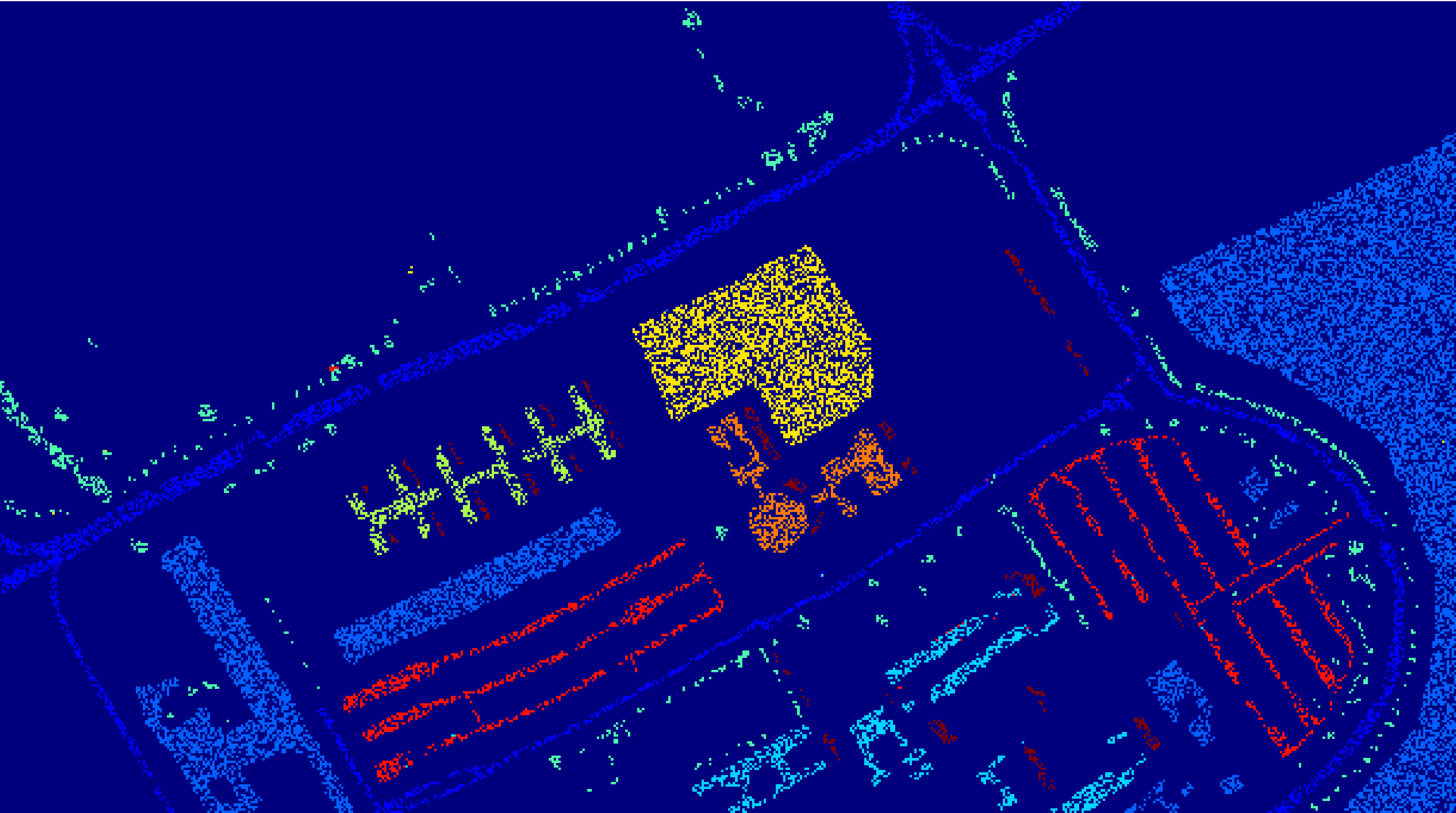}
		\caption{Pavia University} 
		\label{Fig3A}
	\end{subfigure}
	\begin{subfigure}{0.24\textwidth}
		\includegraphics[width=0.99\textwidth]{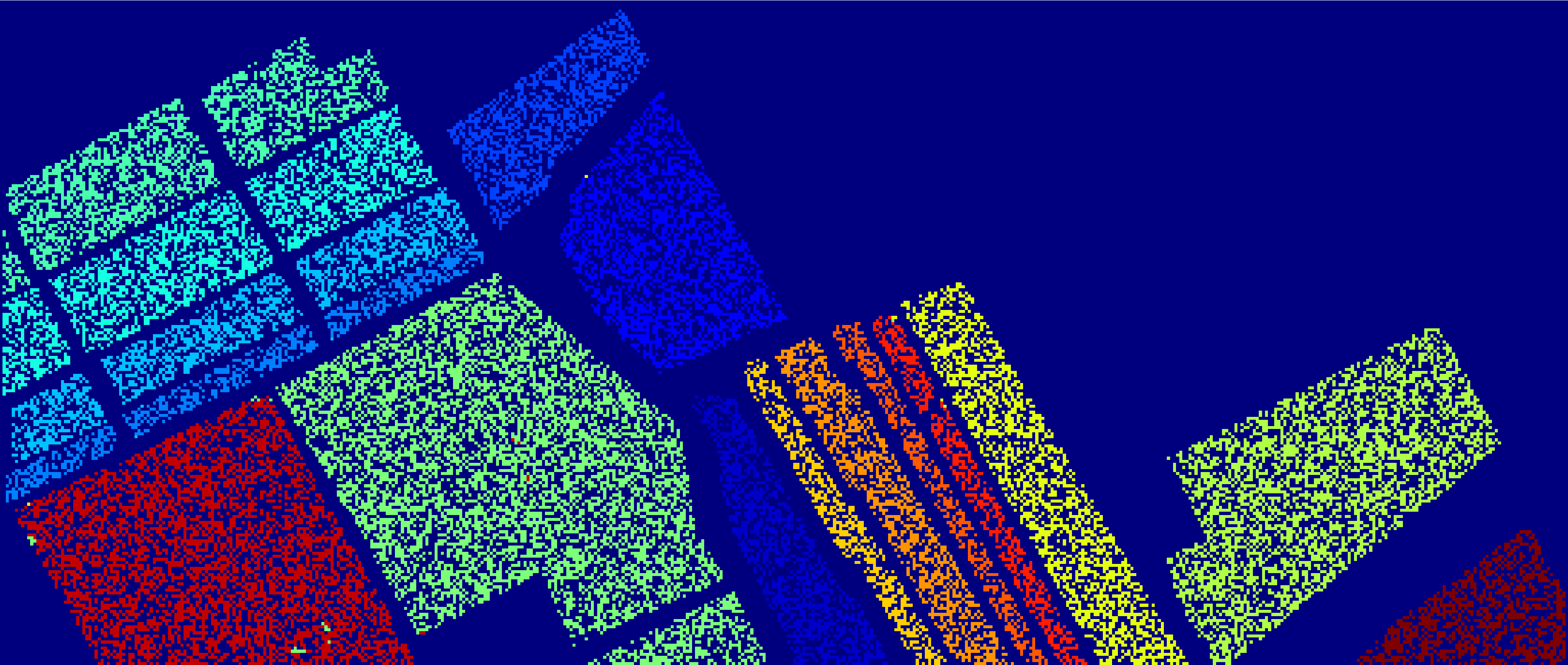}
		\caption{Salinas}
		\label{Fig3B}
	\end{subfigure} 
	\begin{subfigure}{0.49\textwidth}
		\includegraphics[width=0.99\textwidth]{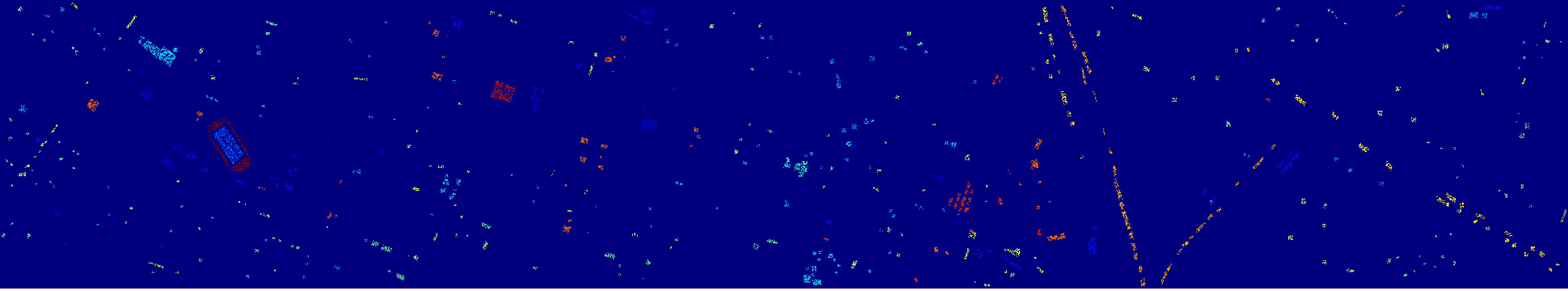}
		\caption{University of Houston}
		\label{Fig3C}
	\end{subfigure}
\caption{The PyFormer model achieves kappa accuracies of 99.73\%, 99.84\%, and 98.01\% on the Disjoint Training, Validation, and Test samples for the Pavia University (PU), Salinas (SA), and University of Houston (UH) datasets, respectively.}
\label{Fig3}
\end{figure}
\begin{table}[!hbt]
\centering
\caption{Performance on different patch sizes across different datasets}
\label{Tab3}
\begin{tabular}{cccccccc} \hline
\multirow{2}{*}{\textbf{Datasets}} & \multirow{2}{*}{\textbf{Metrics}} & \multicolumn{5}{c}{\textbf{Different Patch Size}} \\ \cline{3-7} 
 &  & $2\times 2$ & $4\times 4$ & $6\times 6$ & $8 \times 8$ & $10\times 10$ \\ \hline
\multirow{4}{*}{PU} & OA & 95.88 & 98.72 & \textbf{99.45} & 99.22 & 96.28 \\ 
 & AA & 94.62 & 98.61 & 99.0 & \textbf{99.25} & 95.71 \\ 
 & KA & 94.53 & 98.3 & \textbf{99.27} & 98.97 & 95.05 \\ 
 & F1 & 95 & 98.55 & 99.0 & \textbf{99.33} & 94.88 \\ \hline
\multirow{4}{*}{SA} & OA & 94.5 & 98.53 & \textbf{99.29}  & 98.46 & 99.45\\ 
 & AA & 96.26 & 99.44 & \textbf{99.71} & 98.81 & 99.51 \\ 
 & KA & 93.88 &98.36 & 99.21 & 98.28 & \textbf{99.38} \\
 & F1 & 95.87 & 99.5 & \textbf{99.68} & 98.93 & \textbf{99.68} \\ \hline
 \multirow{4}{*}{HU} & OA & 88.36 &  88.24 & 97.13 & \textbf{98.19} & 90.19\\ 
 & AA & 86.86 & 84.34 & 96.46 & \textbf{97.28} & 89.03 \\ 
 & KA & 87.41 & 87.26 & 96.9 & \textbf{98.05} & 89.03 \\
 & F1 & 81.68 & 79.33 & 90.56 & \textbf{91.62} & 83.81\\ \hline
\end{tabular}
\end{table}
\begin{table}[!hbt]
\centering
\caption{Performance on different Number of Heads across different datasets}
\label{Tab4}
\begin{tabular}{cccccccc} \hline
\multirow{2}{*}{\textbf{Datasets}} & \multirow{2}{*}{\textbf{Metrics}} & \multicolumn{5}{c}{\textbf{Different Number of Heads}} \\ \cline{3-7} 
 &  & 2 & 4 & 6 & 8 & 10 \\ \hline
\multirow{4}{*}{PU} & OA & 99.33 & 95.93 & 98.38 & 99.11 & \textbf{99.36} \\ 
 & AA & \textbf{98.83} & 95.99 & 96.89 & 98.42 & 98.61 \\ 
 & KA & 99.12 & 94.65 & 97.86 & 98.82 & \textbf{99.15} \\ 
 & F1 & \textbf{98.88} & 95.55 & 97.4 & 98.66 & 98.77 \\ \hline
\multirow{4}{*}{SA} & OA & 98.67 & 99.5 & 99.33  & 99.33 & \textbf{99.61} \\ 
 & AA & 98.13 &  99.61 & 98.98 & 99.56 & \textbf{99.66} \\ 
 & KA & 98.52 & 99.44 & 99.25 & 99.26 & \textbf{99.57} \\ 
 & F1 & 97.62 & 99.56 & 99.12 & 99.56 & \textbf{99.75} \\ \hline
 \multirow{4}{*}{HU} & OA & 97.49 & 97.67 & \textbf{97.82} & 84.47 & 96.77\\ 
 & AA & 96.8 & 96.57 & \textbf{97.32} & 85.92 & 96.39 \\ 
 & KA & 97.29 & 97.48 & \textbf{97.65} & 83.25 & 96.51 \\ 
 & F1 & 96.87 & 97.26 & \textbf{97.56} & 84.4 & 96.53\\ \hline
\end{tabular}
\end{table}
\begin{table}[!hbt]
\centering
\caption{Performance on different Number of Layers across different datasets}
\label{Tab5}
\begin{tabular}{cccccccc} \hline
\multirow{2}{*}{\textbf{Datasets}} & \multirow{2}{*}{\textbf{Metrics}} & \multicolumn{5}{c}{\textbf{Different Number of Layers}} \\ \cline{3-7} 
 &  & 2 & 4 & 6 & 8 & 10 \\ \hline
\multirow{4}{*}{PU} & OA & 98.79 & 99.37 & \textbf{99.44} & 99.11 & 97.94 \\ 
 & AA & 98.22 & 98.83 & \textbf{99.19} & 99.0 & 95.43 \\ 
 & KA & 98.4 & 99.17 & \textbf{99.38} & 99.26 & 97.27 \\ 
 & F1 & 98.11 & 99 & \textbf{99.33} & 99.22 & 96.33\\ \hline
\multirow{4}{*}{SA} & OA & \textbf{99.47} & 99.07 & 98.89  & 99.33 & 99.27 \\ 
 & AA & 99.68 &  99.53 & \textbf{99.87} & 99.24 & 98.63 \\ 
 & KA & 99.42 & 98.86 & \textbf{99.65} & 98.76& 98.07 \\ 
 & F1 & 99.81 & 99.68 & \textbf{99.87} & 97.31 & 98.87 \\ \hline
 \multirow{4}{*}{HU} & OA & 97.69 & \textbf{98.1} & 97.63 & 97.28 &97.81\\ 
 & AA & 96.55 & \textbf{97.47} & 97.03 & 96.50 & 97.13 \\ 
 & KA & 97.5 & \textbf{97.95} & 97.44 & 97.06 & 97.64 \\ 
 & F1 & 97& \textbf{97.75} & 97.4 & 96.6 & 97.26\\ \hline
\end{tabular}
\end{table}

Initially, we examine four critical factors impacting the model's performance: patch sizes and training samples as shown in Tables \ref{Tab2} and \ref{Tab3}, and the Number of heads and layers in the Transformer model as shown in Tables \ref{Tab4} and \ref{Tab5}. These factors are pivotal for performance optimization. Ensuring an adequately sized training set covering diverse spectral signatures and representative samples from each class is crucial. Increasing the number of labeled training samples offers the model more diverse examples, enhancing learning and generalization capabilities as shown in Table \ref{Tab2}. Sufficient samples enhance model performance, especially in accommodating variations within classes. An imbalanced distribution of training samples among classes can bias models towards dominant classes, compromising generalization. Balancing sample distribution across classes is crucial to mitigate biases and enhance the model's generalization ability across all classes. Moreover, patch size denotes the spatial extent of input patches, crucial for capturing local spatial information and contextual relationships within HSI data. The choice of patch size significantly impacts the model's ability to capture spatial details and contextual dependencies. Larger patch sizes enhance the model's understanding of global spatial relationships, facilitating improved spatial feature extraction and incorporation of broader spectral information. Conversely, smaller patch sizes focus on local details, advantageous for intricate patterns or objects with distinct spatial characteristics and spectral variations as shown in Table \ref{Tab3}.


Tables \ref{Tab6} and \ref{Tab7} provide a summary of OA, AA, and kappa, along with the average for each class. The best performance in each row is highlighted in bold. For comparison, we have chosen several SOTA models, Transformer \cite{vaswani2017attention}, Spectralformer \cite{hong2021spectralformer}, HiT \cite{9766028}, CSiT \cite{9926105}, and WaveFormer (WF) \cite{10399798}. The transformer architecture utilizes a standard ViT \cite{dosovitskiy2021image} that has been adapted for pixel-wise HSI input, consisting of five encoder blocks. CSiT is evaluated without a Cross-Spectral Attention Fusion module. All results are obtained using the specified parameter configurations from the original papers to facilitate direct comparison.

\begin{table}[!hbt]
    \centering
    \caption{\textbf{Pavia University:} PyFormer is compared against other SOTA models. All models are evaluated with training/validation/test samples distributed as 5\%/5\%/90\%, respectively.}
    \resizebox{\columnwidth}{!}{\begin{tabular}{ccccccc}
        \hline
        \textbf{Class} & SF \cite{hong2021spectralformer} & ViT \cite{vaswani2017attention} & WF \cite{10399798} & CSiT\cite{9926105} & HiT \cite{9766028} & \textbf{PyFormer} \\ \hline
        
        Asphalt & 92.67\% & 95.67\% & \textbf{96.21\%} & 93.84\% & 95.21\% &  95.82\% \\
        Meadows & 92.79\% & 88.37\% & 99.33\% & 95.23\% & 92.54\% & \textbf{99.47\%} \\
        Gravel & \textbf{90.60\%} & 73.71\% & 81.31\% & 88.79\% & 81.18\% & 85.61\% \\
        Trees & 98.15\% & 98.03\% & 96.77\% & 96.19\% & 97.21\% & \textbf{98.79\%} \\
        Painted & 98.28\% & 99.01\% & \textbf{100\%} & 99.18\% & \textbf{100\%} & \textbf{100\%} \\
        Soil & 93.29\% & 89.26\% & 91.55\% & 91.99\% & 91.93\% & \textbf{97.18\%} \\
        Bitumen & 83.01\% & 79.40\% & 89.55\% & 92.06\% & 92.75\% & \textbf{92.88\%}  \\
        Bricks & 84.50\% & 85.54\% & 89.34\% & 82.25\% & 87.59\% & \textbf{89.97\%} \\
        Shadows & \textbf{99.77\%} & 99.65\% & 96.47\% & 99.19\% & 99.47\% & 95.61\% \\ \hline 
        \textbf{OA} & 92.30\% & 89.32\% & 95.66\% & 93.35\% & 91.35\% & \textbf{96.28\%} \\ \hline 
        \textbf{AA} & 88.86\% & 87.39\% & 93.39\% & 90.48\% & 85.07\% & \textbf{93.81\%} \\ \hline 
        \textbf{$\kappa$} & 89.66\% & 86.60\% & 94.22\% & 91.13\% & 88.94\% &  \textbf{95.07\%}\\ \hline
    \end{tabular}}
    \label{Tab6}
\end{table}
\begin{table}[!hbt]
    \centering
    \caption{\textbf{University of Houston:} PyFormer is compared against other SOTA models. All models are evaluated with training/validation/test samples distributed as 10\%/10\%/90\%, respectively.}
    
    \resizebox{\columnwidth}{!}{\begin{tabular}{c|cccccc}\hline 
    
        \textbf{Class} & SF \cite{hong2021spectralformer} & ViT \cite{vaswani2017attention} & WF \cite{10399798} & CSiT \cite{9926105} & HiT \cite{9766028} & \textbf{PyFormer} \\ \hline
        
        Healthy grass & 93.31\% & 90.28\% & 98.90\% & 93.39\% & 97.26\% & \textbf{98.98\%} \\
        Stressed grass & 97.81\% & 98.21\% & 97.60\% & 99.54\% & 97.29\% & \textbf{99.63\%} \\
        Synthetic grass & \textbf{100\%} & 96.90\% & 99.82\% & \textbf{100\%} & 98.74\% & 99.71\% \\
        Trees & \textbf{100\%} & 100\% & 99.19\% & 98.09\% & 95.78\% & 99.48\% \\
        Soil & 98.16\% & 96.89\% & 99.79\% & 97.70\% & 98.41\% & \textbf{100\%} \\
        Water & 100\% & 98.21\% & 98.07\% & 100\% & 91.36\% & 96.74\%  \\
        Residential & 87.83\% & 82.82\% & 90.64\% & 90.96\% & 94.60\% & \textbf{97.35\%}  \\
        Commercial & 85.91\% & 79.83\% & 97.38\% & 89.18\% & 91.82\% & 96.83\% \\
        Road & 75.33\% & 76.88\% & 97.40\% & 90.62\% & 92.39\% & \textbf{97.36\%}  \\
        Highway & 82.52\% & 80.31\% & 97.75\% & 93.22\% & 90.61\% &  \textbf{99.18\%}\\
        Railway & 79.19\% & 83.17\% & 96.76\% & 87.91\% & 89.09\% & \textbf{98.84\%}  \\
        Parking Lot 1 & 72.76\% & 69.42\% & \textbf{97.56\%} & 83.15\% & 94.18\% & 95.72 \\
        Parking Lot 2 & 79.49\% & 63.08\% & 65.33\% & \textbf{84.11\%} & 82.51\% &  78.39\%\\
        Tennis Court & 93.90\% & 90.36\% & 98.83\% & 97.21\% & 91.55\% & \textbf{100\%} \\
        Running Track & 97.50\% & 94.40\% & 99.05\% & \textbf{100\%} & 96.72\% & \textbf{100\%} \\ \hline  
        \textbf{OA} & 88.45\% & 86.45\% & 96.54\% & 93.09\% & 93.06\% & \textbf{97.36\%} \\ \hline 
        \textbf{AA} & 87.81\% & 86.60\% & 95.60\% & 92.06\% & 86.61\% & \textbf{95.97\%} \\ \hline 
        \textbf{$\kappa$} & 87.50\% & 85.35\% & 96.26\% & 92.53\% & 92.50\% &  \textbf{97.14\%} \\ \hline
    \end{tabular}}
    \label{Tab7}
\end{table}

The detailed results of the aforementioned models can be found in Tables \ref{Tab6} and \ref{Tab7}. In summary, the proposed PyFormer model exhibits outstanding performance, surpassing SOTA ViT-based models across various evaluation metrics, including OA, AA, and $\kappa$ coefficient. A comprehensive analysis of the quantitative results indicates that PyFormer consistently achieves superior performance across different categories, demonstrating significant improvements in accuracy, as illustrated in the Tables. Notably, while the performance gaps are relatively small in the PU dataset due to the abundance of samples, the UH dataset presents a considerable challenge for modeling. For instance, when evaluating the challenging UH dataset, PyFormer outperforms the baseline ViT by more than 7\% and exceeds SF by approximately 4\%. Moreover, the AA achieved by PyFormer surpasses that of both ViT and SpectralFormer by margins of around 5\%, highlighting the potential effectiveness of spatial-spectral feature extraction. In comparison with the most recent spatial-spectral Transformer and CSiT models, PyFormer consistently delivers promising results, demonstrating its proficiency in both spectral and spectral-spatial feature extraction tasks. It is noteworthy that while HiT excels in identifying land-cover classes with spectral-spatial information, PyFormer approaches similar levels of performance. In conclusion, these findings underscore the robustness and effectiveness of the PyFormer, particularly in scenarios where the extraction of spatial-spectral information is crucial, especially considering the limited availability of training samples.

\section{Conclusions}

This paper introduces PyFormer, a novel approach that leverages the strengths of Pyramid and Vision Transformer for Hyperspectral Image Classification (HSIC). By extracting multi-scale spatial-spectral features using Pyramid and integrating them into a transformer encoder, PyFormer can effectively capture both local texture patterns and global contextual relationships within a single, end-to-end trainable model. A key innovation is the incorporation of Pyramid convolutions within the transformer's attention mechanism, facilitating enhanced integration of spectral and structural information. Extensive experiments demonstrate that PyFormer achieves SOTA performance, particularly on challenging datasets with limited training data. In addition to superior classification accuracy, PyFormer exhibits robustness and generalizability, showing promise for addressing real-world problems. Future research could explore techniques such as self-supervised pre-training and network optimizations to further enhance PyFormer's performance, especially in scenarios with limited data availability.

\bibliographystyle{IEEEtran}
\bibliography{IEEEabrv,Sam}
\end{document}